\documentclass{article}
\usepackage{microtype}
\usepackage{graphicx}
\usepackage{subfigure}
\usepackage{abstract}
\usepackage{authblk}

\usepackage{amsmath}
\usepackage{amssymb}
\usepackage{mathtools}
\usepackage{amsthm}
\usepackage{xcolor}
\usepackage{algorithm}
\usepackage{algpseudocode}

\newcommand{\diag}[1]{\mathrm{diag}(#1)}
\newcommand{\ones}{\mathbf{1}}
\newcommand{\eye}{\mathbf{I}}
\usepackage[capitalize,noabbrev]{cleveref}
\usepackage{xcolor}

\theoremstyle{plain}

\theoremstyle{definition}

\theoremstyle{remark}

\title{Beyond kNN: Adaptive, Sparse Neighborhood Graphs via Optimal Transport}
\author[1]{Tetsuya Matsumoto}
\author[2]{Stephen Zhang}
\author[1]{Geoffrey Schiebinger}
\affil[1]{Department of Mathematics, University of British Columbia}
\affil[2]{University of Melbourne}

\begin{document}

\maketitle 

\begin{abstract}
Nearest neighbour graphs are widely used to capture the geometry or topology of a dataset. One of the most common strategies to construct such a graph is based on selecting a fixed number $k$ of nearest neighbours (kNN) for each point. However, the kNN heuristic may become inappropriate when sampling density or noise level varies across datasets. Strategies that try to get around this typically introduce additional parameters that need to be tuned. We propose a simple approach to construct an adaptive neighbourhood graph from a single parameter, based on quadratically regularised optimal transport. Our numerical experiments show that graphs constructed in this manner perform favourably in unsupervised and semi-supervised learning applications.
\end{abstract}

\section{Introduction}

The $k$-nearest neighbour ($k$-NN) graph is a staple of modern data science, enjoying near ubiquitous use as an initial step for extracting low-dimensional geometric or topological information from point cloud data. Given a collection of points $\{ x_i \}_{i = 1}^N$ and a distance function $d(x, y) \ge 0$, a graph is constructed in which each point $x_i$ is connected to the $k$ points that are closest to it as measured by $d$. In many applications, data are generated by sampling from some underlying low-dimensional smooth manifold $\mathcal{M} \subset \mathbb{R}^d$. In this scenario, the set of $k$-nearest neighbours of a point $x_i$ can be thought of as its local neighbourhood along the manifold, and the distances correspond to local distances along the manifold. Thus, neighbourhood graph construction comprises a common initial step that underpins most manifold learning methods, including Laplacian eigenmaps \cite{belkin2003laplacian}, diffusion maps \cite{coifman2006diffusion}, ISOMAP \cite{tenenbaum2000global} and UMAP \cite{mcinnes2018umap}. These tools are used in diverse areas of data science from image processing \cite{cayton2005algorithms}, to demography \cite{hedefalk2020social}, to gene expression data \cite{van2018recovering}.

In many applications, data are sampled non-uniformly from $\mathcal{M}$ and may be subject to noise. Nearest-neighbour graphs are known to struggle in these settings, since each point $x_i$ has a fixed neighbourhood size of $k$ neighbours. In order to account for non-uniform sampling, the neighbourhood size $k$ should instead be allowed to vary across the dataset -- in regions of low sampling density a smaller choice of $k$ may prevent spurious long-range connections. Various density-aware approaches have been developed to address this shortcoming, however these methods typically introduce additional parameters parameters that need tuning \cite{balsubramani2019adaptive, van2018recovering} or some form of density estimation \cite{coifman2005geometric}. 

In this paper, we present optimal transport (OT) as a flexible approach for constructing sparse local neighbourhood graphs that automatically adapts to variable sampling density and noise. Each point is assigned a unit of mass which may then be distributed to other points, forming the neighbourhood. This approach is related to recent work connecting entropy-regularised optimal transport to the Laplace-Beltrami operator in manifold learning: \cite{marshall2019manifold} establish that a bistochastic normalisation of the graph Laplacian can be constructed using entropy-regularized OT, and \cite{landa2021doubly} show that this normalization improves robustness to heteroskedastic noise. However, entropic regularization has the drawback that the resulting kernel matrix is dense, representing a fully connected graph. We demonstrate that using instead a \emph{quadratic} regularizer for the optimal transport problem \cite{lorenz2021quadratically} produces a \emph{sparse} neighbourhood graph that can be interpreted similarly to a $k$-NN graph.

The rest of this paper is organized as follows. In Section \ref{sec:Method} we motivate the use of optimal transport, and its quadratic regularization (quadratic OT) in particular, for constructing neighbourhood graphs. We investigate in Section \ref{sec:Results} applications of this approach in several settings including manifold learning, semi-supervised learning, and denoising single-cell expression profiles. 

\section{Method}\label{sec:Method}

\begin{figure*}
    \centering
  \includegraphics[width=0.75\textwidth]{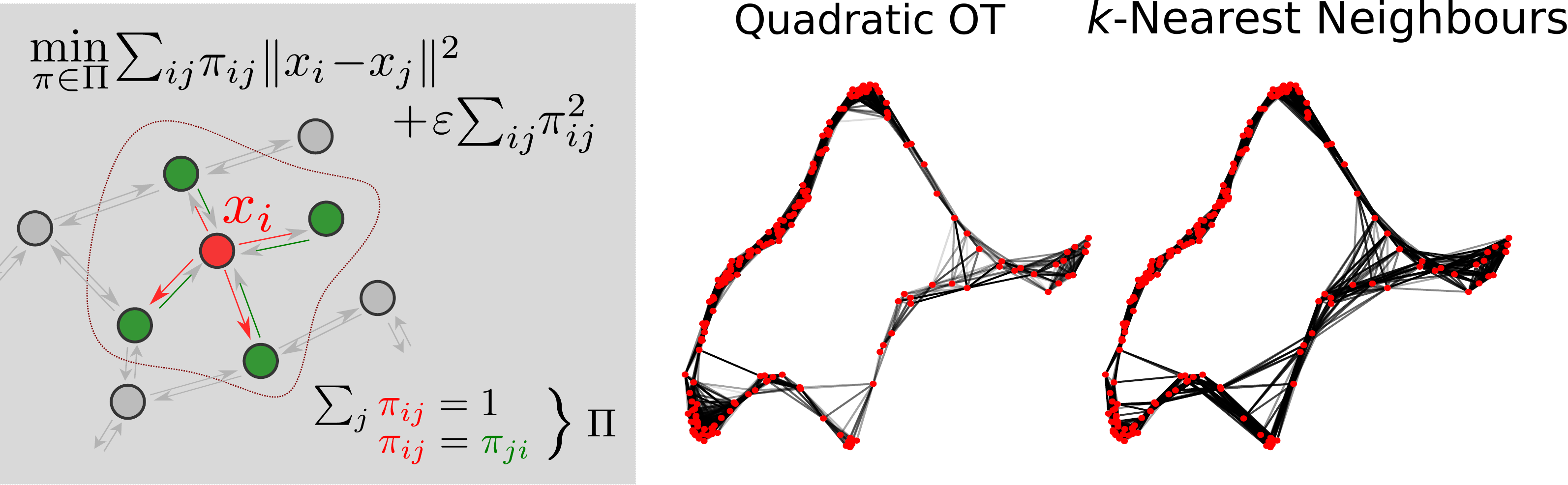}
  \caption{(Left) Neighbourhood graph construction by quadratic OT. The outlined region indicates the ``local neighbourhood'' of the point $x_i$, i.e. the set of points $x_i$ exchanges mass with. (Right) Comparison of graphs (connections shown in black) constructed with quadratic OT and $k$-nearest neighbours, from samples (red) from a closed 1D curve embedded in 2D. Note that $k$-NN forms a graph that has more spurious edges in low-density regions.}
\end{figure*}

The approach we adopt follows \cite{landa2021doubly} in defining a mass transport problem on the set $\{ x_i \}$. We assign to each point a unit of mass and allow for redistribution of mass to other points, subject to the constraint that the mass outgoing must equal the mass received. We prescribe the penalty incurred by sending a unit mass from $x_i$ to $x_j$ to be $c_{ij} = d(x_i, x_j)^p$. Let $\pi_{ij} \ge 0$ denote the amount of mass sent from $x_i$ to $x_j$: we will assume that $\pi_{ij} = \pi_{ji}$ (symmetry) and that $\sum_i \pi_{ij} = 1$ (mass balance). Thus, the matrix $\pi$ can be treated as the adjacency matrix of a undirected weighted graph. The total penalty incurred by the \emph{transport plan} $\pi$ is $\sum_{i,j} \pi_{ij} c_{ij}$. Mass balance and symmetry assumptions can be written as the constraint set $\Pi = \{  \pi : \pi \ge 0, \pi = \pi^\top, \pi \ones = \ones\}$. One may then consider the minimisation problem
\begin{align}
    \min_{\pi \in \Pi} \sum_{i, j} \pi_{ij} c_{ij}.
\end{align}
This problem admits a trivial solution $\pi = \eye$ where each point sends its unit mass to itself (in which case the ``neighbourhood'' shrinks to a point), and thus reveals no information about local geometry. 
In order to encourage mass to be distributed away from a point and onto its ``neighbours'', we consider regularising the problem with a convex functional $\Omega$ and parameter $\varepsilon > 0$. 
\begin{align}
    \min_{\pi \in \Pi} \sum_{i, j} \pi_{ij} c_{ij} + \varepsilon \Omega(\pi). \label{eq:reg_prob}
\end{align}
We refer the reader to \cite{blondel2018smooth, dessein2018regularized} for an introduction to optimal transport with general convex regularisation. Note that something similar could also be achieved by modifying the cost function so that points cannot send mass to themselves (i.e. $c_{ii} = \infty$). 

Previous work \cite{landa2021doubly, marshall2019manifold} considered the negative entropy $\Omega(\pi) = \sum_{i, j} \pi_{ij} \log \pi_{ij}$, which results in the symmetric entropy-regularised optimal transport problem. The nature of the entropy term ensures that the optimal $\pi$ is strictly positive, i.e. one obtains a complete graph with local neighbourhood information encoded in the edge weights rather than presence of edges. Indeed, the optimal solution is characterised as a doubly-stochastic normalisation of the Gaussian kernel: $\pi_{ij} = e^{u_i/\varepsilon} e^{u_j/\varepsilon} e^{-c_{ij}/\varepsilon}$. 

Instead, we consider the quadratic regulariser $\Omega(\pi) = \sum_{i, j} \pi_{ij}^2$, which yields a symmetric \emph{quadratically}-regularised optimal transport problem \cite{lorenz2021quadratically}. 
As in the entropic case, the quadratic regulariser pushes the optimal $\pi$ away from the identity. However, it is a well-known empirical result that the resulting matrices $\pi$ that are \emph{sparse} in the sense that most entries are identically zero. This can be seen from the characterisation of the optimal solution: $\pi_{ij} = \frac{1}{\varepsilon} \max\{0, u_i + u_j - c_{ij} \}$. Since the optimal $\pi$ is sparse, the local neighbourhood of each point $x_i$ can be regarded as $\{ x_j : \pi_{ij} > 0 \}$. The matrix $\pi$ therefore encodes a graph that can be interpreted similarly as $k$-NN graphs. The key difference here is that the parameter $\varepsilon$ for the quadratic OT problem controls only the \emph{overall} level of sparsity, without a fixed number of edges per node being prescribed as in the case of $k$-NN. \cite{lorenz2021quadratically} present a numerical method for solving the quadratically regularised problem \eqref{eq:reg_prob}, which can be specialised to the symmetric setting as we show in Algorithm \ref{alg:lorenz_symmetric} in the appendix. 

The sparse nature of the quadratically regularized optimal transport plan can be intuited from the characterisation provided by \cite{dessein2018regularized}, who show that the solution of \eqref{eq:reg_prob} is equivalent to 
\begin{align*}
    \min_{\pi \in \Pi} B_\Omega(\pi | \xi), \quad \xi_{ij} = \nabla \Omega^{-1} (-c_{ij}/\varepsilon),
\end{align*}
where $B_\Omega(\pi | \xi) = \Omega(\pi) - \Omega(\xi) - \langle \pi - \xi, \nabla \Omega(\xi) \rangle$ is the Bregman divergence induced by $\Omega$. 
When $\Omega(\pi) = \frac{1}{2} \| \pi \|^2$, one has $B_\Omega(\pi | \xi) = \frac{1}{2} \| \pi - \xi \|^2$. A straightforward computation shows that $\xi = -c/\varepsilon$, and so quadratic OT can be seen as finding a finding a Euclidean projection of $-c/\varepsilon$ (a matrix with negative entries) onto $\Pi$ (a set of transport plans constrained to be non-negative). Thus, one may expect that many entries of $\pi$ will be identically zero.

\section{Results}\label{sec:Results}

\subsection{Manifold learning with heteroskedastic noise}\label{sec:ManifoldLearning}

We investigate first the application of quadratic OT to the setting of a low-dimensional manifold embedded in a high-dimensional ambient space, with noise of varying intensity. This is the setting of \lq heteroskedastic noise\rq where \cite{landa2021doubly} proposed entropic OT. This setting occurs in single-cell RNA-sequencing data, where the noise level of each cell's expression profile is determined by the number of reads sequenced for that cell.

\begin{figure}
    \centering
    \includegraphics[width = 0.75\linewidth]{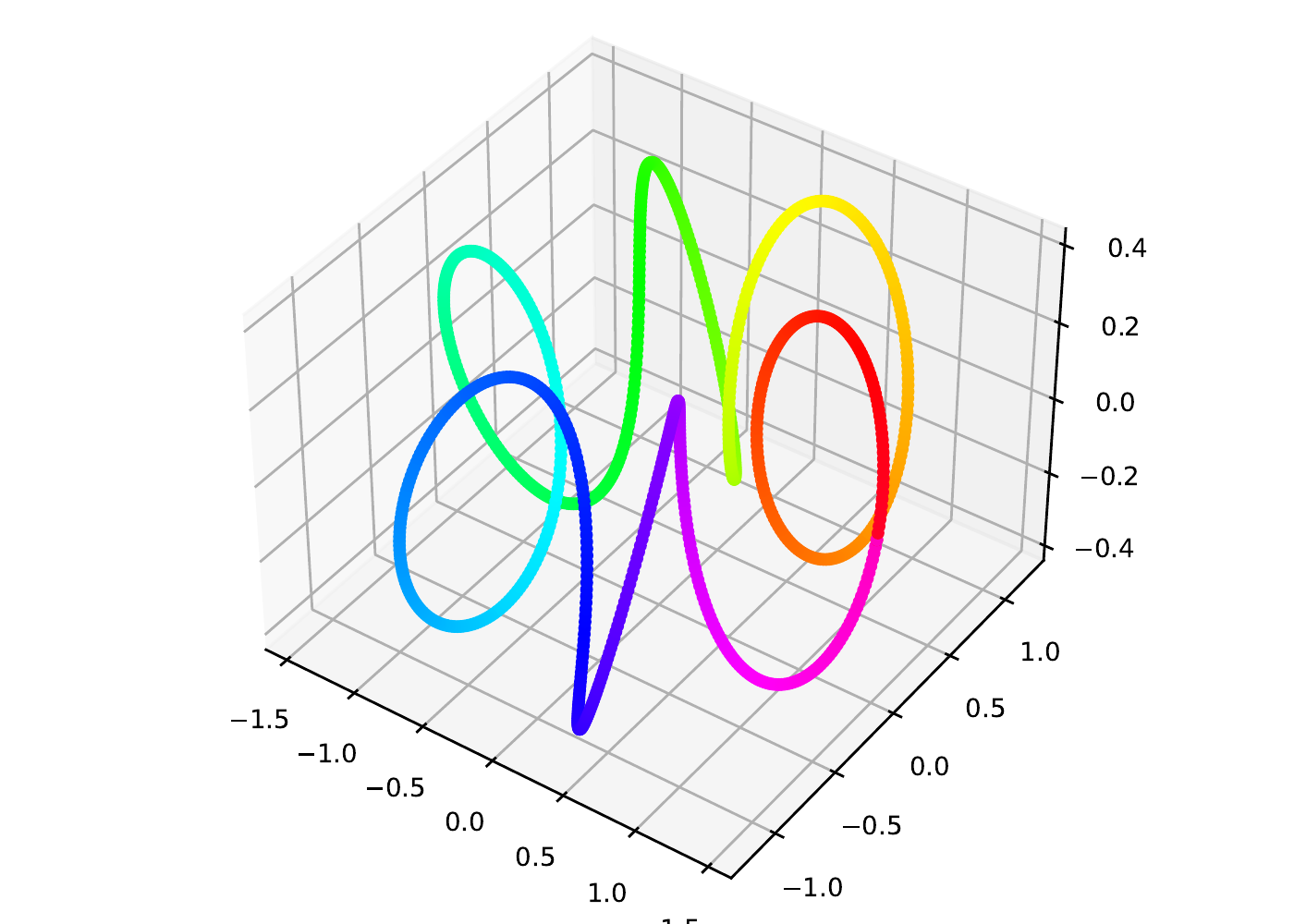}
    \caption{Closed spiral dataset in 3 dimensions.}
    \label{fig:spiral_dataset}
\end{figure}

We generated the simulated dataset in Figure \ref{fig:spiral_dataset} with $10^3$ points $\{ x_i \}$ evenly spaced with respect to arc-length along the spiral parameterized by 
\begin{align*}
    \begin{bmatrix} x(t) \\ y(t) \\ z(t) \end{bmatrix} = \begin{bmatrix}\cos(t) (0.5\cos(6t)+1) \\ \sin(t) (0.4\cos(6t)+1) \\ 0.4\sin(6t)\end{bmatrix}, \quad t \in [0, 2\pi].
\end{align*}

\begin{figure}[h]
    \centering
    \subfigure[]{\includegraphics[width = \linewidth]{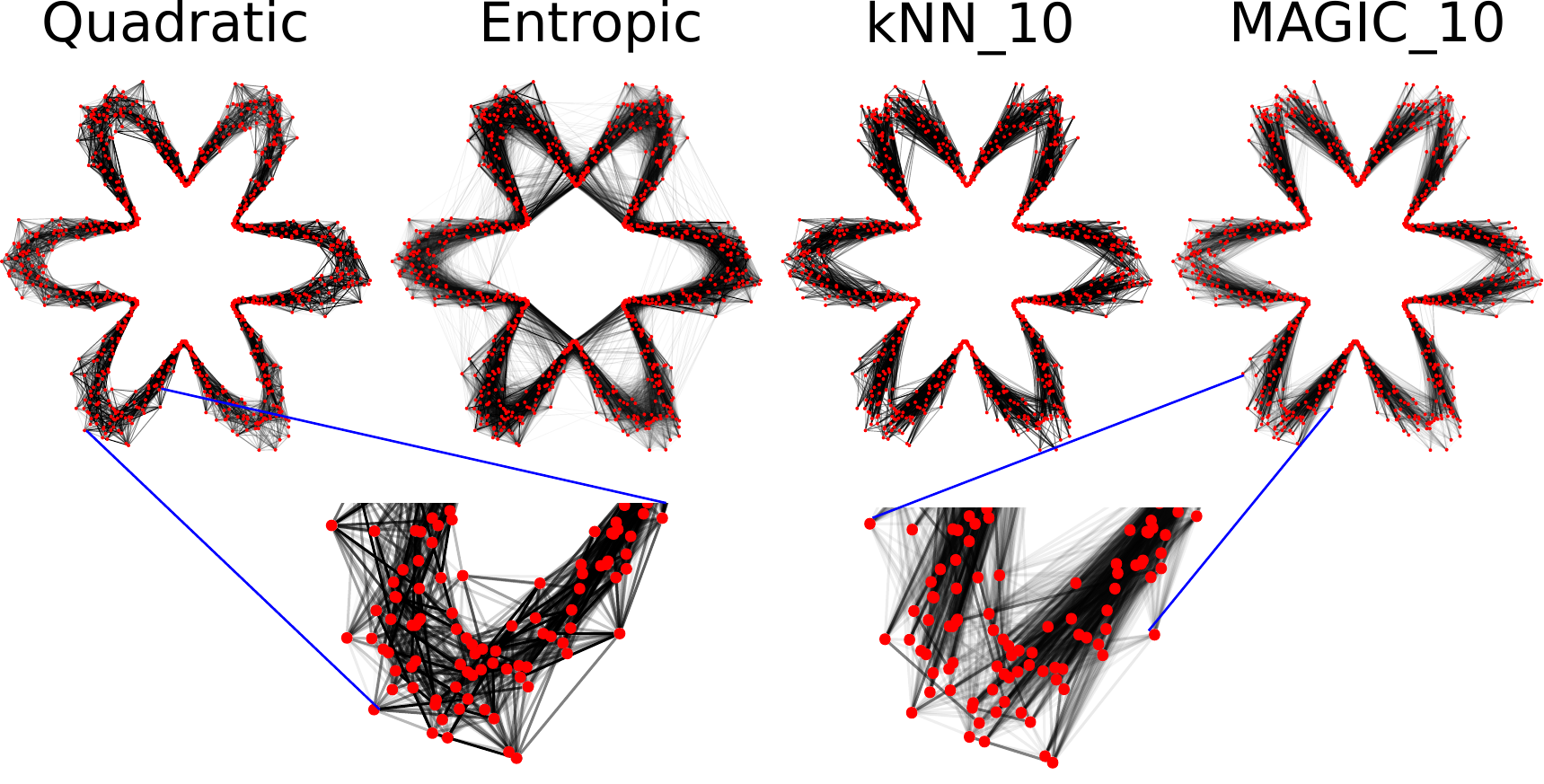}} \\ 
    \subfigure[]{\includegraphics[width = \linewidth]{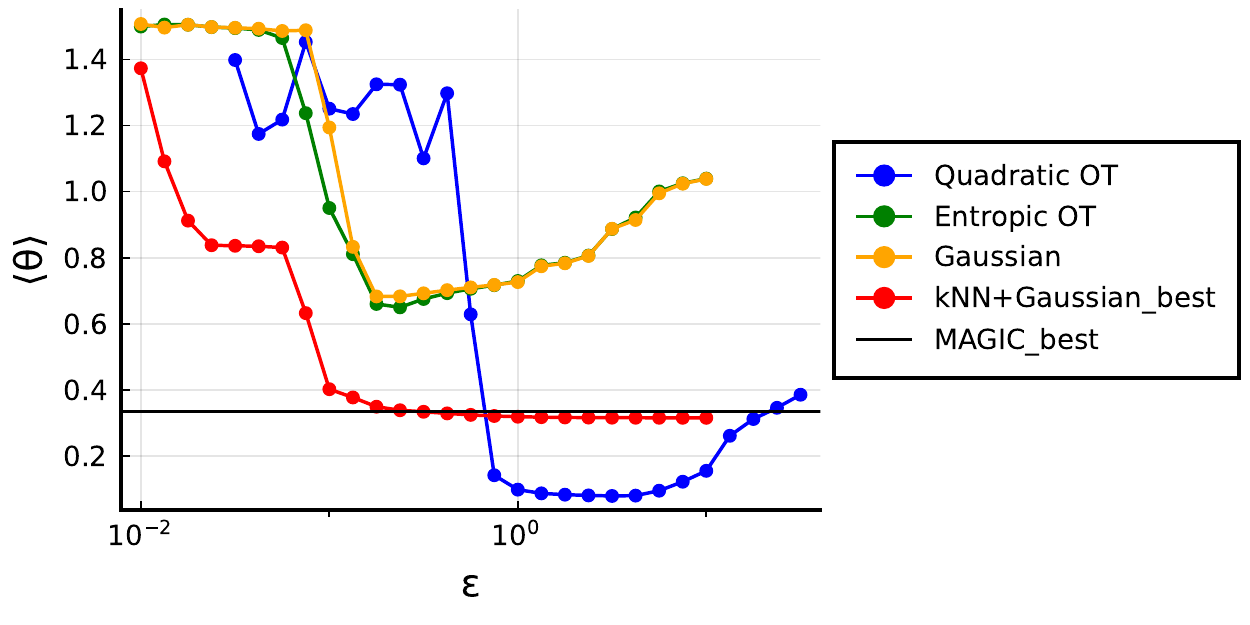}}
    \caption{(a) Neighbourhood graphs constructed by various methods for parameter settings corresponding to best eigenfunction embeddings. (b) Mean error in the eigenspaces spanned by top 10 non-trivial eigenvectors, measured in terms of principal angles. For $k$-NN and MAGIC, best results over $k \in \{ 5, \ldots, 25 \}$ are shown. %
    }
    \label{fig:spiral}
\end{figure}

We then generated a random matrix $R \in \mathbb{R}^{100 \times 3}$ such that $R^\top R = I$, i.e. the columns are orthogonal. Each point $x_i \in \mathbb{R}^3$ was then linearly embedded in the 100-dimensional ambient space with additive noise via $\hat{x}_i = R x_i + \eta_i$. The noise model we use is 
\begin{align*}
    \eta_i = \frac{Z_i}{\| Z_i \|} \rho(\theta), \quad \rho(\theta) = 0.05 + 0.95\dfrac{1 + \cos(6\theta)}{2}.
\end{align*}
{where $Z_i$ follows the standard Gaussian distribution.}
In order to eliminate the effect of scaling on the effect of the spread parameter $\varepsilon$, we normalised all input data so that $N^{-2} \sum_{i, j = 1}^N \| \hat{x}_i - \hat{x}_j \|^2 = 1$. 
Given the noisy high-dimensional dataset $\{ \hat{x}_i \}$, we constructed local neighbourhood graphs using a variety of methods and a range of parameter values:  
\begin{itemize}
    \item Quadratic OT: $\varepsilon \in [10^{-1.5}, 10^{1.5}]$ (logarithmic spacing)
    \item Entropic OT \cite{landa2021doubly}: $\varepsilon \in [10^{-2}, 10^{1}]$ (logarithmic spacing)
    \item $k$-NN with Gaussian edge weights \cite{belkin2003laplacian}: $k$-NN graph was constructed with $k \in \{ 5, 10, 15, 20, 25\}$ and edge weights $\exp(-\| x_i - x_j\|^2/\varepsilon)$, with $\varepsilon$ in same range as entropic OT. 
    \item Gaussian kernel \cite{belkin2003laplacian}: fully connected graph with edge weights $\exp(-\| x_i - x_j \|^2/\varepsilon)$, with $\varepsilon$ in same range as entropic OT.
    \item MAGIC \cite{van2018recovering}: $k \in \{ 5, 10, 15, 20, 25\}$.
\end{itemize}
In Figure \ref{fig:spiral}(a), we show the noisy data projected onto the first two principal components alongside representative graphs constructed using each method. This 2D embedding has a star-like shape with lobes that correspond to regions of large magnitudes of noise interspersed by regions of low noise which appear to be pinched inwards. 

Qualitative differences in the graphs are immediately visible -- entropic OT forms edges between ``lobes'' of the spiral, corresponding to the dense nature of the graph. As visible in the zoomed region of high noise level, MAGIC (which relies on $k$-NN) forms few edges \emph{between} points in the high noise region, with most edges branching away into the low-noise region. On the other hand, quadratic OT forms appreciably stronger connections between points in the high-noise region. 

\begin{figure}[h]
    \centering
    \includegraphics[width = \linewidth]{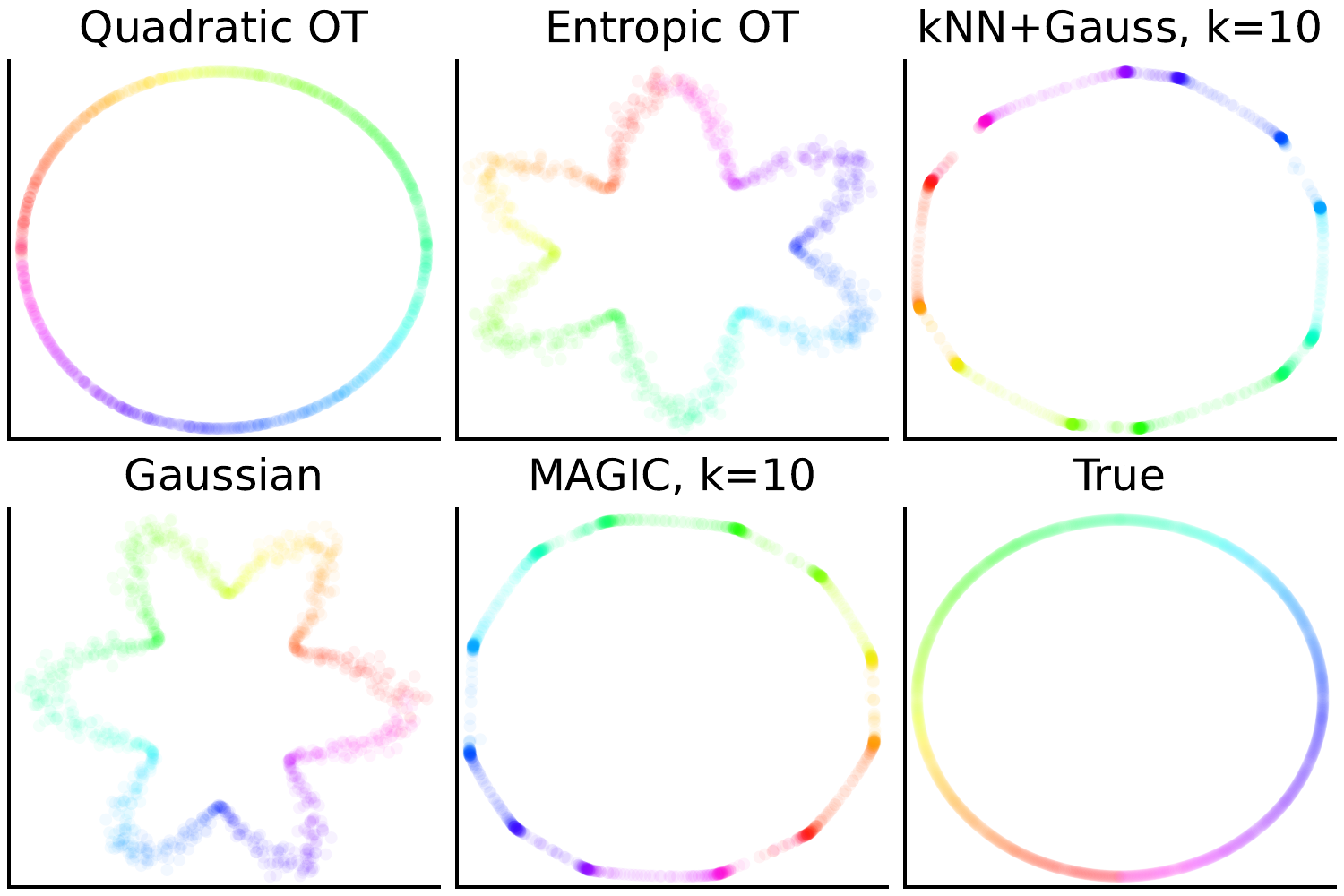}
    \caption{Best eigenvector embeddings found, shown in the top 2 nontrivial eigenvectors. Quadratic OT produces an embedding closest to the \lq true\rq embedding, based on the noiseless data.}
    \label{fig:spiral_embeddings}
\end{figure}

In Figures \ref{fig:spiral}(b) and \ref{fig:spiral_embeddings}, we investigate the spectral properties of neighbourhood graphs are closely related to underlying geometry, and provide a convenient means of quantitative evaluation. To investigate this, we computed eigenmap embeddings from each of the constructed weighted graphs following \cite{belkin2003laplacian}. Given a weighted adjacency matrix $W$ computed by a method of choice, $W$ was first random-walk normalized to form $\overline{W}$:
\begin{align*}
    \overline{W}_{ij} = \frac{W_{ij}}{\sum_j W_{ij}}.
\end{align*}
The spectral decomposition of $\overline{W}$ was then computed, yielding eigenvalues $\lambda_N \leq \lambda_{N-1} \leq \ldots \leq \lambda_{1} = 1$, and corresponding eigenvectors $v_N, \ldots, v_1$. The top eigenvector is trivial and so the $\ell$-dimensional \emph{eigenmap embedding} for $\ell = 1, \ldots, N-1$ is then the mapping 
\begin{align*}
    x_i \mapsto (v_{2}(x_i), v_{3}(x_i), \ldots, v_{\ell + 1}(x_i)).
\end{align*}
As a reference, a graph was constructed from the clean 3-dimensional data  $\{x_i\}$ (as opposed to $\{\hat x_i\}$) using a $k$-NN with $k = 10$ and a Gaussian kernel with $\varepsilon = 0.025$. From this, we computed reference eigenvectors $\{ v_i^\mathrm{ref} \}_i$. To measure of how well the eigenvectors from the noisy high-dimensional data correspond to those computed from the clean low-dimensional data, we computed the mean principal angle \cite{knyazev2002principal} between the subspaces spanned by the top 10 non-trivial eigenvectors, i.e. $\operatorname{span}\{v_2, \ldots, v_{11}\}$ and $\operatorname{span}\{ v_2^\mathrm{ref}, \ldots, v_{11}^\mathrm{ref} \}$.  

We summarise these results in Figure \ref{fig:spiral}(b). From this it is clear that the graph constructions obtained using entropic OT and a simple Gaussian kernel perform the worst overall -- there are no parameter settings for which these methods perform well. Intuitively, this can be ascribed to the fact that the graphs are dense and try to model neighbourhoods using decaying edge weights. This effect can be seen also in Figure \ref{fig:spiral}(a), where the graph constructed by entropic OT has visible edges between ``lobes'' of the spiral. 

Methods that respect sparsity of the neighbourhood graph do significantly better, on the other hand. We note that $k$-NN and MAGIC perform comparably, with the key distinction that MAGIC has the capability of automatically adapting the effective bandwidth across regions of the dataset, while $k$-NN uses a universal bandwidth $\varepsilon$ that needs to be tuned. We observe that quadratic OT outperforms both these methods. We believe that this is a result of the sparsity level and edge weights being not directly prescribed but instead controlled indirectly by the regularisation parameter $\varepsilon$.

Finally, we show in Figure \ref{fig:spiral_embeddings} the best embeddings (in terms of the eigenspace error) in the top two non-trivial eigenvectors. As one would expect for eigenfunctions of a closed 1D curve with uniform density \cite{coifman2005geometric}, the reference embedding is a perfect circle. Both entropic OT and the simple Gaussian kernel produce embeddings that are still star-shaped, making no improvement beyond the PCA embeddings displayed in \ref{fig:spiral}(a). $k$-NN with Gaussian weights and MAGIC produce circular embeddings, however points are irregularly distributed, concentrating at several points. Only quadratic OT produces a near-perfect circle.  

\subsection{Application to semi-supervised learning}\label{sec:SSL}

Next we turn to investigating an application of our graph construction method in which the downstream steps are more involved, namely semi-supervised learning (SSL). The problem setting here is that we are provided with data $\{ x_i \}_{i = 1}^N$, of which a small subset $L \subset \{1, \ldots, N\}$ are annotated with labels $\{ \ell_i \}_{i \in L}$. The semi-supervised learning problem is then to infer a set of labels $\{ \hat{\ell}_i \}$ that (approximately) preserves the known labels $\hat{\ell}_i \approx \ell_i, i \in L$ and also estimates the unknown labels $\{\hat{\ell}_i\}_{i \notin L}$. These inference procedures typically operate under the heuristic that similarly labelled points should be in close proximity, necessitating the construction of a neighbourhood graph. 

While there are many schemes for solving the SSL problem, we will focus on the classical method Learning Local and Global Consistency (LLGC) of \cite{zhou2004learning}, for which we now provide a brief description. LLGC aims to optimise label assignments to balance between matching the known labels (fitting) and consistency of inferred labels with the data geometry (consistency). 

Let $P$ be a matrix of dimensions $N \times C$, with $P_{ij} = \mathbf{1}\{ x_i \text{ has label } j\}$. Thus, the rows of $P$ are one-hot encodings for labelling known points, and unlabelled points correspond to all-zeros. The aim is to infer a matrix $Q \in \mathbb{R}^{N \times C}_{\ge 0}$ describing the labels of all points (both labelled and unlabelled), such that 
\begin{align*}
    \hat{\ell}_i = \operatorname{argmax}_j Q_{ij}. 
\end{align*}
For a neighbourhood graph with row-stochastic weighted adjacency matrix $W_{ij}$ and regularisation parameter $\mu > 0$, the LLGC optimisation problem is written as 
\begin{align}
    \min_{Q} \underbrace{\sum_{i, j = 1}^{N} W_{ij} \left\| Q_i - Q_j \right\|^2}_{\text{consistency}} + \mu \underbrace{\sum_{i = 1}^N \| Q_i - P_i \|^2}_{\text{fitting}}.
    \label{eq:llgc}
\end{align}

We generated data from a multi-armed spiral similar to that used in \cite{budninskiy2020laplacian}. We chose 10 arms started at angles $\theta = \pi/5, 2\pi/5, \ldots, 2\pi$, and for each initial angle $\theta$ the spiral arm was parameterised as 
\begin{align}\label{eqn:spiral}
    r(t; \theta) = 
    \begin{bmatrix}
        \cos(\theta) & -\sin(\theta)\\
        \sin(\theta) & \cos(\theta)
    \end{bmatrix}
    \begin{bmatrix}
        t \cos(t)\\  t\sin(t)
    \end{bmatrix},  
    t \in [t_0, t_1].
\end{align}
We set $t_0 = 1, t_1 = 5$ and sampled 150 points along each arm to be uniformly spaced with respect to the arc-length of the branch. Following the approach of Section \ref{sec:ManifoldLearning}, the clean data $x_i$ were then embedded into 100 dimensions and noise was added following $\hat{x}_i = Rx_i + \eta_i$ with $\eta_i = (1 - \sin(3t_i)^4) Z_i/\|Z_i\|$. For each branch, 2.5\% of points were selected at random to be labeled.

We then solved \eqref{eq:llgc} with $W$ taken to be the adjacency matrix constructed either by quadratic OT, or using a $k$-NN graph with (row-normalized) Gaussian weights. Each method was re-run with different parameter settings: $\varepsilon \in [10^{-2}, 10^{1}]$ (logarithmically spaced) for quadratic OT, and $\varepsilon \in [10^{-3}, 10^{0}]$ (logarithmically spaced), $k \in [1, 50]$ (linearly spaced). 

\begin{figure}
    \centering
    \subfigure[]{\includegraphics[width = 0.9\linewidth]{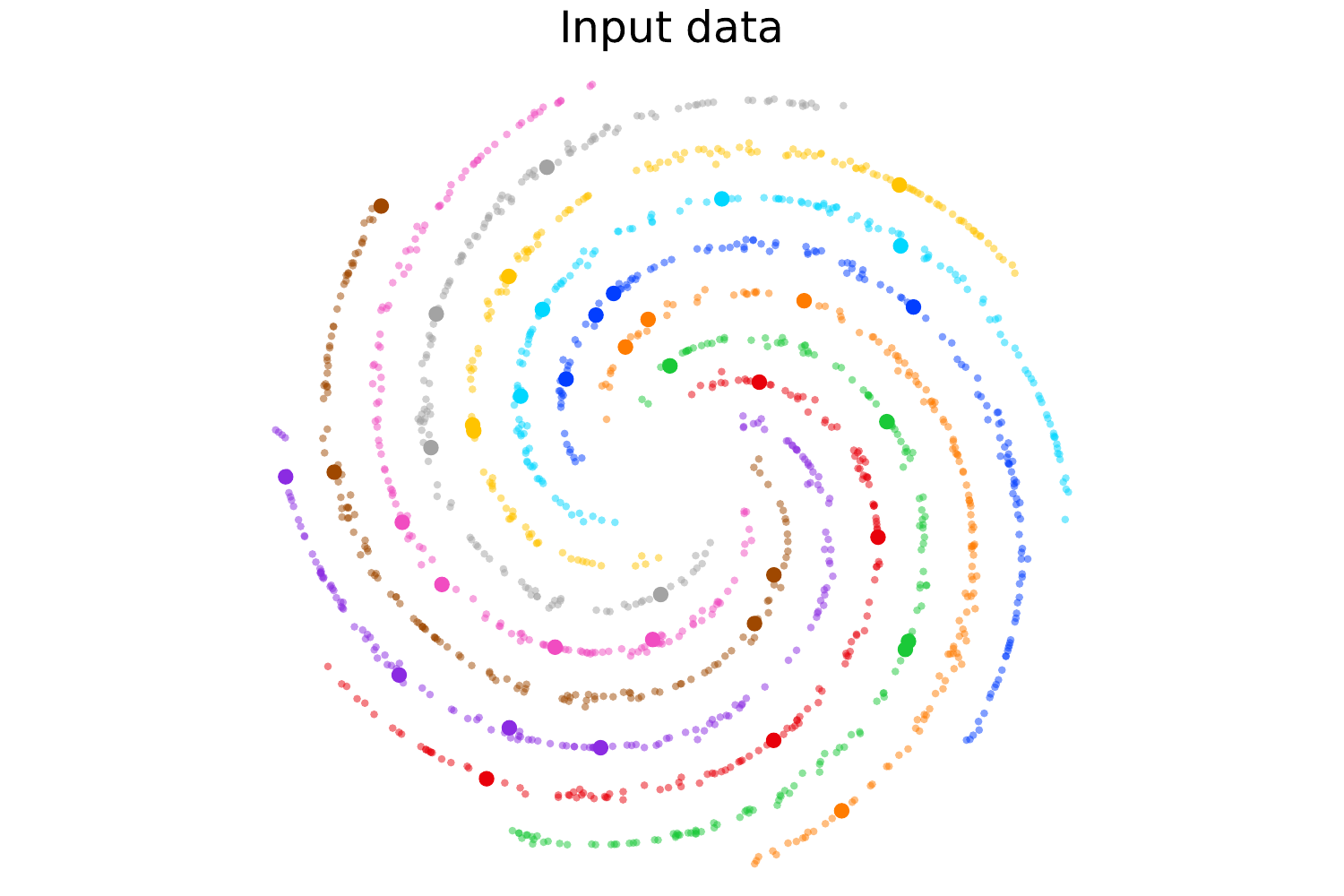}}
    \subfigure[]{\includegraphics[width = \linewidth]{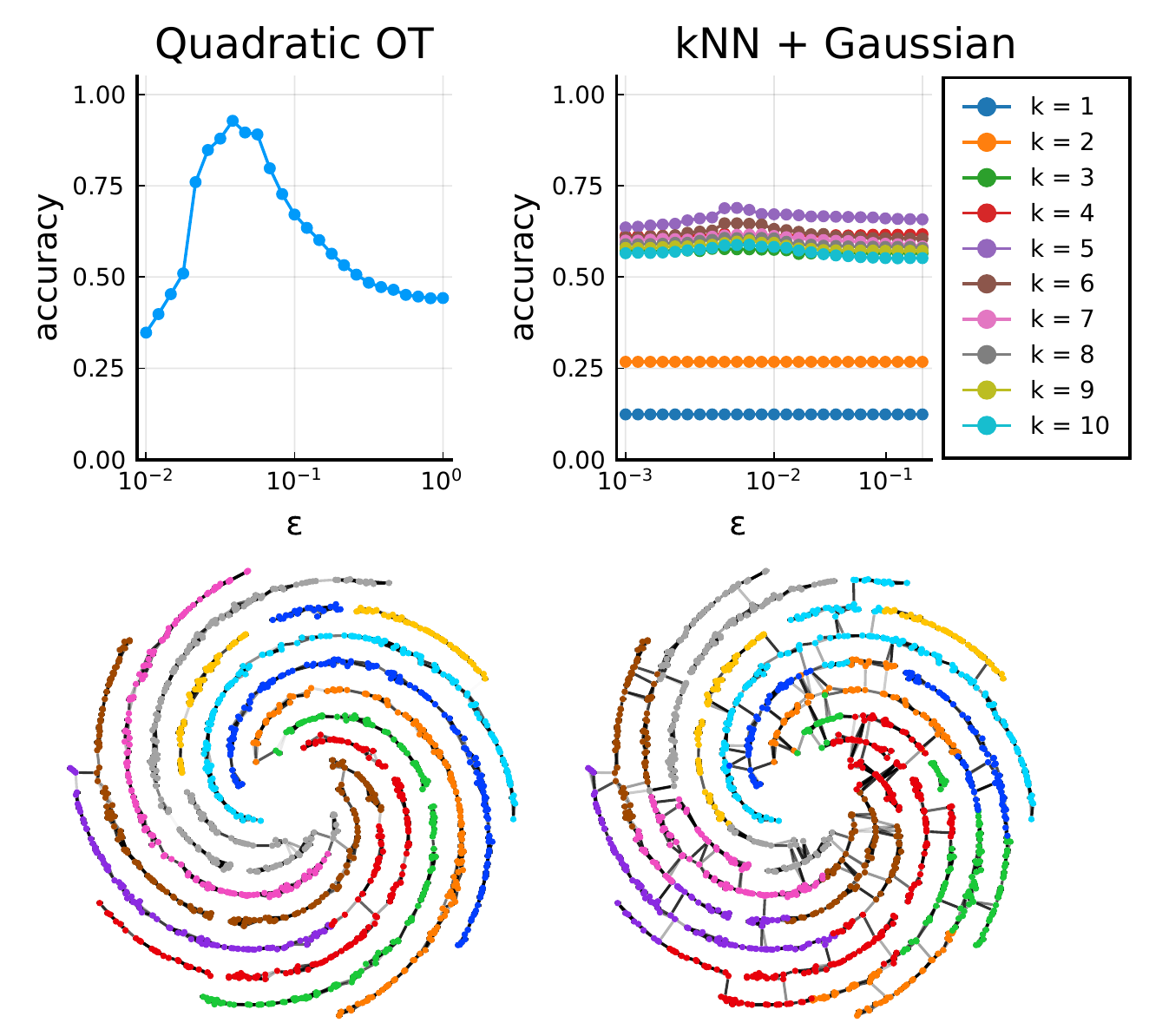}}
    \caption{(a) Data from 10-armed spiral with 150 points per branch and 2.5\% labelled points, in high-dimension with varying noise. Here we show the 2D PCA embedding. (b) Top: Accuracy for quadratic OT and $k$-NN with Gaussian edge weights as a function of parameter values Bottom: Inferred labels (shown by color) and neighbourhood graphs (black edges) computed from each method, for the optimal parameter values.}
    \label{fig:llgc_density}
\end{figure}

Results are summarised in Figure \ref{fig:llgc_density}: from this it is apparent that in the presence of high-dimensional noise, quadratic OT significantly outperforms $k$-NN with Gaussian weights. In Figure \ref{fig:llgc_density}(b), we show both inferred labels and neighbourhood graphs for the best results achieved by each method. Quadratic OT appears to have mostly correctly assigned labels and the neighbourhood graph is well-adapted to the underlying geometry, with very few edges between spiral arms. On the other hand, $k$-NN with Gaussian weights produces many incorrect labels. We note that this coincides with the presence of many edges connecting adjacent spiral arms.

\begin{figure}
    \centering
    \includegraphics[width = \linewidth]{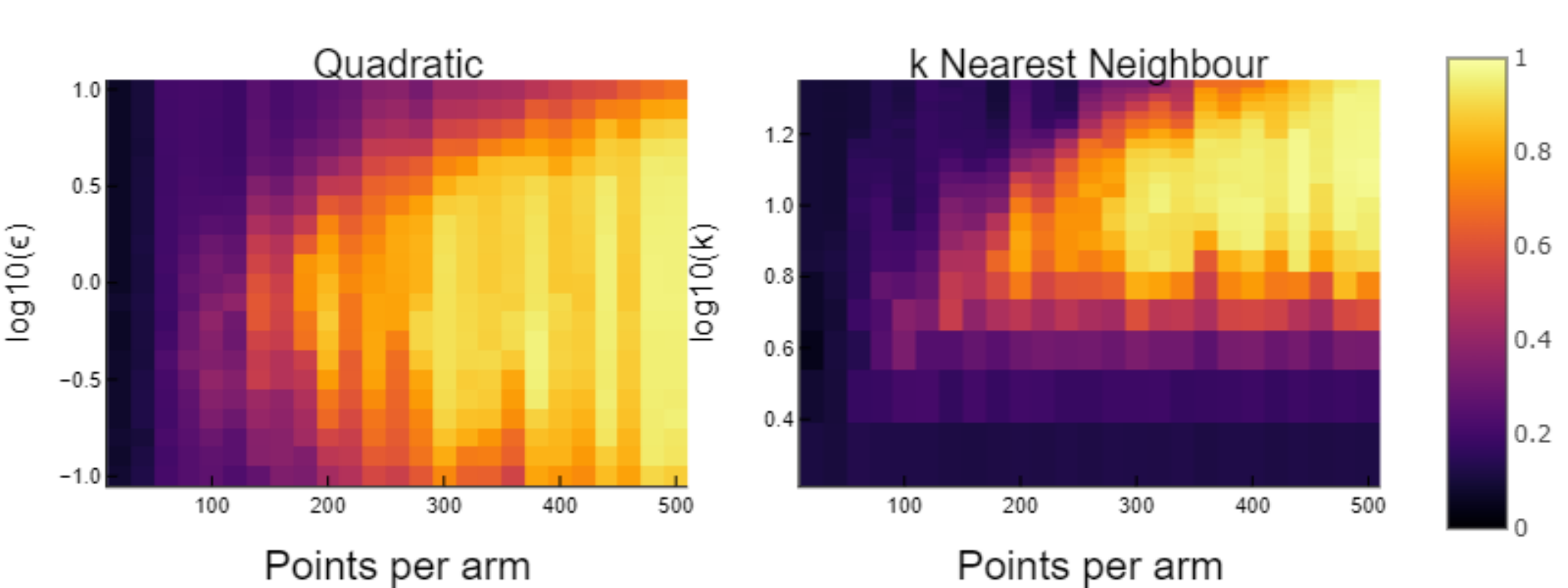}
    \caption{Heatmap of accuracy over number of points per arm and parameters. Note that for kNN the the optimal value of $k$ depends on the number of points per arm, but for quadratic OT a single value  ($\varepsilon = 1$) is optimal.}
    \label{fig:arm_vs_param}
\end{figure}

In Figure \ref{fig:arm_vs_param}, we explore the performance of both quadratic OT and kNN as the number of points along the spiral arms increases, and therefore the density increases.
Here we generated the 10-armed spiral data following the same parametrization \eqref{eqn:spiral} with randomly sampled points along $t\in[0.5, 5.0]$. We labeled 1\% of the data. 
The heat map summarizes accuracy for varying parameter ranges as the number of points per arm increases. From this it is clear that quadratic OT achieves comparable performance across problem scales for a wide range of $\varepsilon$, while the optimal value of $k$ for kNN changes as we vary the number of points per arm. {This can be attributed to the fact that quadratic OT balances the mass going into as well as the mass coming out of each point, instead of simply connecting each point to a fixed number $k$ of neighbours.}

\subsection{Application to  Single Cell RNA-sequencing}\label{sec:MAGIC}

Single cell measurement technologies are revolutionizing the biological sciences by providing high-dimensional measurements of cell state. 
By sequencing the RNA molecules within individual cells, single-cell RNA-sequencing (scRNA-seq) quantifies the gene expression profiles of these cells (i.e. which genes are \lq off\rq~and which are \lq on\rq~and at what level). Recent droplet-based  technologies have tremendously increased the throughput of scRNA-seq, so that large numbers of cells can be profiled, and have brought \lq Big Data\rq~to biology. 

Sequencing more cells comes at the cost of sequencing fewer molecules of RNA per cell, and so scRNA-seq data is notoriously noisy. Lowly expressed genes may not be detected (i.e. they are \lq dropped out\rq), which makes the analysis of gene-gene relationships difficult. 
This has motivated several groups to develop tools for denoising single-cell expression profiles. 
One of the most popular tools for this is MAGIC (Markov affinity-based graph imputation of cells) developed by \cite{van2018recovering}. 

MAGIC constructs a local neighbourhood graph from an adaptive Gaussian kernel to impute the data. 
For each point $x_i$ in the data, the Gaussian parameter $\sigma_i$ is chosen by $\sigma_i = d(x_i, x^j_i)$ where $x^j_i$ is the $j$-th nearest neighbour of $x_i.$
This allows the kernel to have larger bandwidth in sparse areas and smaller bandwidth in dense areas, and was shown to be crucial in the original paper~\cite{van2018recovering}.
MAGIC then denoises the data by smoothing it according to a diffusion along this graph.   

\begin{figure}[h]
    \centering
    \subfigure[]{
    \includegraphics[width=\linewidth]{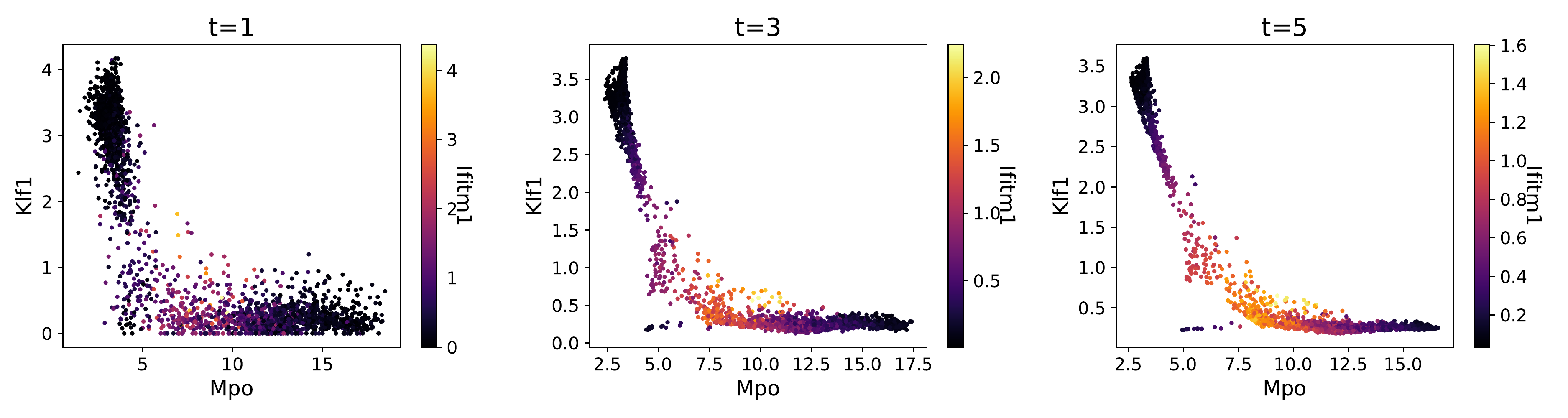}}
    \subfigure[]{
    \includegraphics[width=\linewidth]{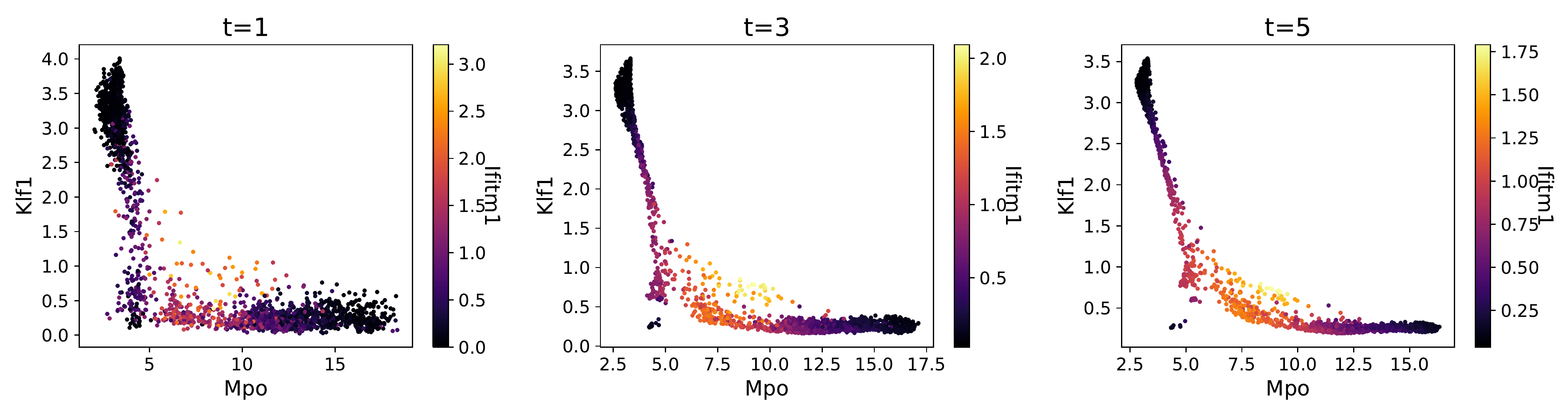}}
    \subfigure[]{
    \includegraphics[width=\linewidth]{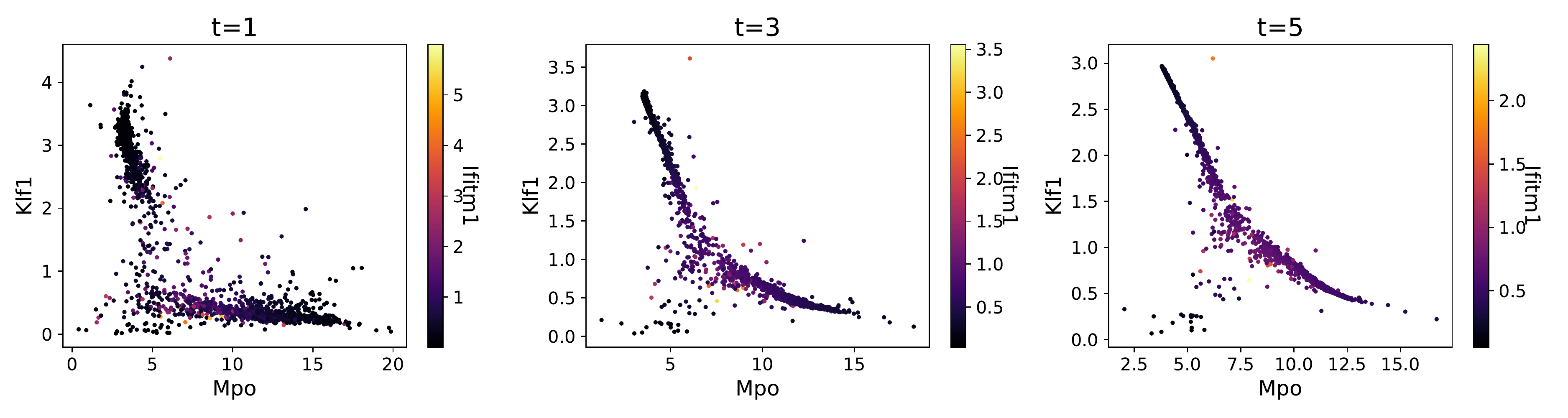}}
    \caption{Progression of MAGIC denoising based on local neighbourhood graphs constructed by (a) adaptive Gaussian kernel as in the original MAGIC formulation, (b) Quadratic OT, and (c) Entropic OT. In each case, we show results for three different levels of the diffusion time-scale parameter $\{t=1, t=3, t=5\}$.}
    \label{fig:cellMAGIC}
\end{figure}

We tested whether quadratic OT could produce a neighbourhood graph without explicitly adjusting for density.
Figure \ref{fig:cellMAGIC} shows the results of denoising a single cell dataset from~\cite{van2018recovering} for neighbourhood graphs constructed by quadratic OT, entropic OT, and the original formulation of MAGIC with the adaptive Gaussian kernel.
We find that quadratic OT performs nearly identical to MAGIC, which confirms that quadratic OT is adaptive to variable density. 
On the other hand, entropic OT fails to smooth points near the centre and loses geometric information that is present on the original.

\section{Discussion}\label{sec:Discussion}

Quadratically regularised optimal transport provides a simple and robust method for constructing weighted neighbourhood graphs with only a single parameter $\varepsilon$. Our numerical experiments provide evidence that, in many geometric machine learning applications, this graph construction can replace the role of $k$-nearest neighbour graphs to achieve similar or superior performance. This can be attributed to the fact that, instead of connecting each point to a fixed number $k$ of neighbours, OT balances the mass coming into and going out of each point. The constructed graph along with its weight is adaptive to its domain which is a major advantage over classical $k$ nearest neighbour graph based method especially on a high dimensional dataset. 

In future work, it will be of interest to study theoretical properties of the linear operator derived from the quadratically regularised optimal transport problem. Analysis of the behavior of this operator in the limit of infinite data may  plausibly arrive at a convergence result to the Laplace-Beltrami operator, a result that is established in the literature for many other common graph constructions \cite{ting2011analysis}. We hypothesise that such theoretical properties of the graph operator may involve a connection between quadratically regularised optimal transport and subdiffusions \cite{lavenant2018dynamical}. 

\bibliography{references}
\bibliographystyle{unsrt}

\newpage
\appendix
\onecolumn
\section{Appendix}

\begin{algorithm}
    \caption{Algorithm 2 of \cite{lorenz2021quadratically}, specialised to symmetric inputs}
    \label{alg:lorenz_symmetric}
    For completeness, we have replicated parts of \cite{lorenz2021quadratically} with the modifications made for the symmetric problem setting. This reduces the size of the linear system \eqref{eq:lorenz_linear_system}
    \begin{algorithmic}
        \State Input: cost matrix $c_{ij}$, regularisation parameter $\varepsilon > 0$, Armijo parameters $\theta, \kappa \in (0, 1)$, conjugate gradient regulariser $\delta = 10^{-5}$. 
        \State Initialize: $u \gets \ones$
        \While{not converged}
            \State $P_{ij} \gets u_i + u_j - c_{ij}$
            \State $\sigma_{ij} = \ones\{ P_{ij} \ge 0 \}$
            \State $\pi_{ij} = \max\{ P_{ij}, 0 \}/\varepsilon$
            \State Solve for $\Delta u$:
            \begin{equation}
                \left( \sigma + \diag{\sigma \ones} + \delta I \right) \Delta u = -\varepsilon (\pi - I)\ones.\label{eq:lorenz_linear_system}
            \end{equation}
            \State Set $t = 1$
            \While{$\Phi(u + t \Delta u) \ge \Phi(u) + t \theta d$} 
                \State $t \gets \kappa t$
            \EndWhile
            \State $u \gets u - t \Delta u$
        \EndWhile 
        \State $\pi_{ij} = \max\{ 0, u_i + u_j - c_{ij}\}/\varepsilon$
    \end{algorithmic}
\end{algorithm}

\end{document}